\documentclass[conference]{IEEEtran}

\usepackage{hyperref}
\usepackage{graphicx}
\usepackage{xcolor}
\usepackage{xspace}
\usepackage{bbold}
\usepackage{flushend}
\usepackage{amsmath,amsthm,mathtools}
\usepackage[caption=false]{subfig}

\newcommand{\DaDN}{\textit{DaDN}\xspace}
\newcommand{\STRL}{\textit{Stripes}\xspace}
\newcommand{\STR}{\textit{STR}\xspace}
\newcommand{\BASE}{\textit{DaDN}\xspace}
\newcommand{\PRA}{\textit{Pragmatic}\xspace}
\newcommand{\PRAS}{\textit{PRA}\xspace}

\begin{document}

\title{Bit-Pragmatic Deep Neural Network Computing}
\date{}

\author{
\IEEEauthorblockN{Jorge Albericio,
Patrick Judd,
Alberto Delm\'as,
Sayeh Sharify,
Andreas Moshovos
}
\IEEEauthorblockA{
Department of Electrical and Computer Engineering\\
University of Toronto\\
\{jorge, juddpatr, a.delmaslascorz, sayeh, moshovos\}@ece.utoronto.ca}
}

\maketitle


\begin{abstract}
\sloppy
We quantify a source of ineffectual computations when processing the multiplications of the convolutional layers in Deep Neural Networks (DNNs) and propose \PRA (\PRAS), an architecture that exploits it improving performance and energy efficiency. 
The source of these ineffectual computations is best understood in the context of conventional multipliers which generate internally multiple \textit{terms}, that is, products of the multiplicand and powers of two, which added together produce the final product~\cite{Wallace64}. At runtime, many of these terms are zero as they are generated when the multiplicand is combined with the zero-bits of the multiplicator. While conventional bit-parallel multipliers calculate all terms in parallel to reduce individual product latency, \PRAS calculates only the non-zero terms using a)~on-the-fly conversion of the multiplicator representation into an explicit list of powers of two, and b)~bit-parallel multplicand/bit-serial multiplicator processing units. 

\PRAS 
exploits two sources of ineffectual computations: 1)~the aforementioned zero product terms which are the result of the \textit{lack of explicitness} in the multiplicator representation, and 2)~the \textit{excess in the representation precision} used for both multiplicants and multiplicators, e.g.,~\cite{judd:reduced}. Measurements demonstrate that for the convolutional layers, a straightforward variant of \PRAS improves performance by 2.6x over the DaDiaNao (DaDN) accelerator~\cite{DaDiannao} and by 1.4x over STR~\cite{Stripes-CAL}. Similarly, \PRAS improves energy efficiency by 28\% and 10\% on average compared to DaDN and STR. An improved cross lane synchronization scheme boosts performance improvements to 3.1x over DaDN. Finally, \PRA benefits persist even with an 8-bit quantized representation~\cite{quantizedBlog}. 
\end{abstract}

\section{Introduction}\label{sec:introduction}
Deep neural networks (DNNs) have become the state-of-the-art technique in many recognition tasks such as object~\cite{RCNN13} and speech recognition~\cite{deep-speech}. 
While DNN's have high computational demands, they are today practical to  deploy given the availability of commodity Graphic Processing Units (GPUs) which can exploit the natural parallelism of DNNs. Yet, the need for even more sophisticated DNNs demands even higher performance and energy efficiency motivating special purpose architectures such as the state-of-the-art DaDianNao (\BASE)~\cite{DaDiannao}.
With power limiting modern high-performance designs, achieving better energy efficiency is essential can enable further advances~\cite{darkSilicon}.

DNNs comprise a pipeline of \textit{layers} where more than 92\% of the processing time is spent in convolutional layers~\cite{DaDiannao}, which this work targets. These layers perform inner products where \textit{neurons} and \textit{synapses} are multiplied in pairs, and where the resulting products are added to produce a single \textit{output neuron}. A typical convolutional layer performs hundreds of inner products, each accepting hundreds to thousands neuron and synapse pairs.

DNN hardware typically uses either 16-bit fixed-point~\cite{DaDiannao} or quantized 8-bit numbers~\cite{quantizedBlog} and bit-parallel compute units. Since the actual precision requirements vary considerably across DNN layers~\cite{judd:reduced}, typical DNN hardware ends up processing an excess of bits when processing these inner products~\cite{Stripes-CAL}. Unless the values processed by a layer need the full value range afforded by the hardware's representation, an excess of bits, some at the most significant bit positions (prefix bits) and some at the least significant positions (suffix bits), need to be set to zero yet do not contribute to the final outcome. With bit-parallel compute units there is no performance benefit in not processing these excess bits. 

Recent work, \STRL (\STR) uses serial-parallel multiplication~\cite{ienne1994bit} to avoid processing these zero prefix and suffix bits~\cite{Stripes-CAL} yielding performance and energy benefits. \STR represents the neurons using pre-specified per layer precisions. Given a neuron $n$ represented in $p$ bits and a synapse $s$ represented in, for example, 16-bits, \STR processes $n$ bit-serially over $p$ cycles, where in each cycle one bit of $n$ is multiplied by $s$ accumulating the result into a running sum. While \STR takes $p$ cycles to compute each product,  it can ideally improve performance by $16/p$ compared to a 16-bit fixed-point bit-parallel hardware by processing $16\times$ more neurons and synapse pairs in parallel. The abundant parallelism of DNN convolutional layers makes this possible.

While \STR avoids processing the ineffectual suffix and tail bits of neurons that are due to the one-size-fits-all representation of conventional bit-parallel hardware, it still processes many ineffectual neuron bits: Any time a zero bit is multiplied by a synapse it adds nothing to the final output neuron. These ineffectual bits are introduced by the conventional positional number representation. If these multiplications could be avoided it would take even less time to calculate each product improving energy and performance. Section~\ref{sec:motivation} shows that in state-of-the-art image classification networks show that 93\% and 69\% of neuron bit and synapse products are ineffectual when using respectively 16-bit fixed-point and 8-bit quantized representations.

This work presents \PRA (\PRAS) a DNN accelerator whose goal is to process only the \textit{essential} (non-zero) bits of the input neurons. \PRAS subsumes \STR not only since a)~it avoids processing non-essential bits regardless of their position, but also as b)~it obviates the need to determine \textit{a priori} the specific precision requirements per layer. \PRAS employs the following four key techniques: 1)~on-the-fly conversion of neurons from a storage representation (e.g.,  conventional positional number or quantized) into an explicit representation of the essential bits only, 2) bit-serial neuron/bit-parallel synapse processing, an idea borrowed from \STR but adapted for the aforementioned representation, 3)~judicious SIMD (single instruction multiple data) lane grouping to maintain wide memory accesses and to avoid fragmenting and enlarging the multi-MB on-chip synapse memories (Sections~\ref{sec:datasupply} and~\ref{sec:prag-synchro}), and 4)~computation re-arrangement (Section~\ref{sec:prag-2stage}) to reduce datapath area. 
All evaluated \PRAS variants maintain wide memory accesses and use highly-parallel SIMD-style (single-instruction multiple-data) computational units. \PRAS introduces an additional dimension upon which software can improve performance and energy efficiency by controlling neuron values judiciously in order to reduce their essential bit content while maintaining accuracy. This work explores such an alternative, where the software explicitly communicates how many prefix and suffix bits to discard after each layer.

Experimental measurements with state-of-the-art DNNs demonstrate that most straightforward \PRAS variant, boosts average performance for the convolutional layers to 2.59x over the state-of-the-art \BASE accelerator compared to the 1.85x performance improvement of \STR alone. \PRA's average energy efficiency is 1.48x over \BASE and its area overhead is 1.35x. Another variant further boosts performance to 3.1x over \BASE at the expense of an additional 0.7\% area. Software guidance accounts for 19\% of these performance benefits.

\section{Motivation}
\label{sec:motivation}
Let us assume a $p$-bit bit-parallel multiplier using a straightforward implementation of the ``Shift and Add'' algorithm where $n\times s $ is calculated as $\sum^p_{i=0}{n_i \cdot (s\ll i)}\), where $n_i$ the i-\textit{th} bit of $n$. The multiplier computes $p$ \textit{terms}, each a product of $s$ and of a bit of $n$, and adds them to produce the final result. The terms and their sum can be calculated concurrently to reduce latency~\cite{Wallace64}. 

\begin{figure}
\centering
\includegraphics[scale=0.8]{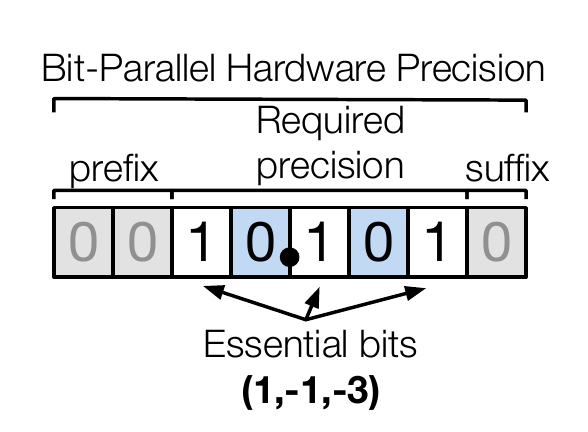} 
\caption{Sources of ineffectual computation with conventional positional representation and fixed-length hardware precision.
}
\label{fig:motivation-excess}
\end{figure}

With such a hardware arrangement there are two sources of ineffectual computations that result from: 1)~an \textit{Excess of Precision} (EoP), and 2)~\textit{Lack of Explicitness} (LoE).
Figure~\ref{fig:motivation-excess} shows an example illustrating these sources with a bit-parallel multiplier using an 8-bit unsigned fixed-point number with 4 fractional and 4 integer bits.
While $10.101_{(2)}$ requires just five bits, our 8-bit bit-parallel multiplier will zero-extend it with two prefix and one suffix bits. This is an example of EoP and is due to the fixed-precision hardware. Two additional ineffectual bits appear at positions 1 and -2 as a result of LoE in the positional number representation. In total, five ineffectual bits will be processed generating five ineffectual terms. 


Our number could be represented with an explicit list of its three constituent powers of 2: (1,-1,-3).  While such a representation may require more bits and thus be undesirable for storage, coupled with the abundant parallelism that is present in DNNs layers, it provides an opportunity to revisit hardware design improving performance and energy efficiency. 

The rest of this section motivates \PRA by: 1)~measuring the fraction of non-zero bits in the neuron stream of state-of-the-art DNNs for three commonly used representations, and 2)~estimating the performance improvement which may be possible by processing only the non-zero neuron bits.



\subsection{Essential Neuron Bit Content}
\label{sec:motivation:essential}
Table~\ref{tab:avg-ones} reports the \textit{essential bit content} of the neuron stream of state-of-the-art DNNs for two commonly used fixed length representations: 1)~16-bit fixed-point of DaDianNao~\cite{DaDiannao}, 
2)~8-bit quantized of Tensorflow  \cite{quantizedBlog}. The essential bit content is the average number of non-zero bits that are 1. Two measurements are presented per representation: over all neuron values (``All''), and over the non-zero neurons (``NZ'') as accelerators that can skip zero neurons for fixed-point representations have been recently proposed~\cite{han_eie:_2016,albericio:cnvlutin}.

When  considering all neurons, the essential bit-content is at most 12.7\% and 38.4\% for the fixed-point and the quantized representations respectively.  The measurements are consistent with the neuron values following a normal distribution centered at 0, and then being filtered by a rectifier linear unit (ReLU) function~\cite{relu}. Even when considering the non-zero neurons the essential bit content remains well below 50\% and as the next section will show, there are many non-zero valued neurons suggesting that the potential exists to improve performance and energy efficiency over approaches that target zero valued neurons. 

These results suggest that a significant number of ineffectual terms are processed with conventional fixed-length hardware. \STRL~\cite{Stripes-CAL}, tackles the excess of precision, exploiting the variability in numerical precision DNNs requirements to increase performance by processing the neurons bit-serially. \PRA's goal is to also exploit the lack of explicitness. As the next section will show, \PRA has the potential to greatly improve performance even when compared to \STRL.

\begin{table}[]
\centering
\footnotesize
\begin{tabular}{|c||r|r|r|r|r|r|}
\hline
& \textbf{Alexnet} & \textbf{NiN}   & \textbf{Google} & \textbf{VGGM}  & \textbf{VGGS}  & \textbf{VGG19} \\ \hline\hline
\multicolumn{7}{|c|}{\textbf{16-bit Fixed-Point}}\\
\hline
All & 7.8\%	&10.4\%	&6.4\%	&5.1\%	&5.7\%	&12.7\% \\
NZ & 18.1\%	&22.1\%	&19.0\%	&16.5\%	&16.7\%	&24.2\% \\
%
\hline
\hline
\multicolumn{7}{|c|}{\textbf{8-bit Quantized}}\\
 \hline
All & 31.4\%	&27.1\%	&26.8\%	&38.4\%	&34.3\%	&16.5\% \\
NZ & 44.3\%	&37.4\%	&42.6\%	&47.4\%	&46.0\%	&29.1\% \\ \hline

\end{tabular}
\caption{Average fraction of non-zero bits per neuron for two fixed-length representations: 16-bit fixed-point, and 8-bit quantized. \textbf{All:} over all neurons. \textbf{NZ:} over non-zero neurons only.}
\label{tab:avg-ones}
\end{table}


\subsection{\PRA's Potential}
\label{sec:motivation:potential}

To estimate \PRAS's  potential, this section compares the number of terms that would be processed by various computing engines for the convolutional layers of state-of-the-art DNNs (see Section~\ref{sec:methodology}) for the two aforementioned baseline neuron representations.

\noindent\textbf{16-bit Fixed-Point Representation:} The following computing engines are considered: 1)~baseline representative of \BASE using 16-bit fixed-point bit-parallel units~\cite{DaDiannao}, 2) a \textit{hypothetical} enhanced baseline ZN, that can skip \textit{all} zero valued neurons, 3)~Cnvlutin (CVN) a practical design that can skip zero value neurons for all but the first layer~\cite{albericio:cnvlutin}, 4)~\STR that avoids EoP (see Table~\ref{tab:best_mixed}, Section~\ref{sec:methodology})~\cite{Stripes-CAL}, 5)~an ideal, software-transparent \PRAS, \PRAS-fp16 that processes only the essential neuron bits, and 6)~an ideal \PRAS, \PRAS-red, where software communicates in advance how many prefix and suffix bits can be zeroed out after each layer (see Section~\ref{sec:software}). 

\begin{figure}
\centering
\includegraphics[width=0.9\linewidth]{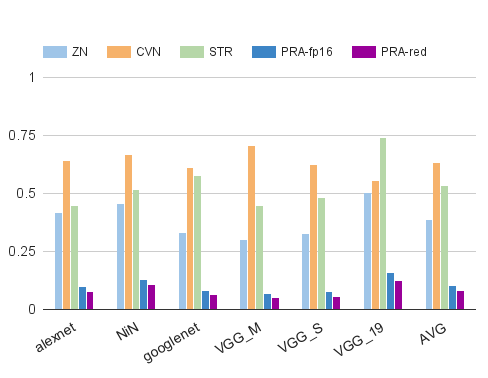}
\caption{Convolutional layer computational demands with a 16-bit fixed-point baseline representation. Lower is better.}
\label{fig:frac-adds}
\end{figure}

Figure~\ref{fig:frac-adds} reports the number of terms normalized over \BASE where each multiplication is accounted for using an equivalent number of terms or equivalently additions: 
16 for \BASE, ZN, and CVN, $p$ for a layer using a precision of $p$ bits for \STR, and the number of essential neuron bits for \PRAS-fp16, and for \PRAS-red.  For example, for $n=10.001_{(2)}$, the number of additions counted would be 16 for \BASE and CVN+, 5 for \STR as it could use a 5-bit fixed-point representation, and 2 for \PRAS-fp16 and \PRAS-red. 
 
On average, \STR reduces the number of terms to 53\% compared to \BASE while skipping just the zero valued neurons could reduce them to 39\% if ZN was practical and to 63\% in practice with CVN. \PRAS-fp16 can ideally reduce the number of additions to just 10\% on average, while with software provided precisions per layer, \PRAS-red reduces the number of additions further to 8\% on average. The potential savings are robust across all DNNs remaining above 87\% for all DNNs with \PRAS-red.


\begin{figure}
\centering
\includegraphics[width=0.9\linewidth]{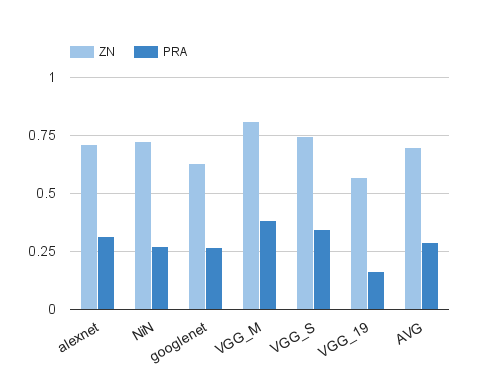}
\caption{Convolution layer computational demands with an 8-bit quantized baseline representation. Lower is better.}
\label{fig:frac-adds_q}
\end{figure}

\noindent\textbf{8-bit Quantized Representation:} Figure~\ref{fig:frac-adds_q} shows the relative number of terms processed for: 1)~a bit-parallel baseline, 2)~an ideal, yet impractical bit-parallel engine that skips all zero neurons, and 3)~\PRAS. In the interest of space and since \PRAS subsumes \STR and CVN they are not considered.  
\PRA's potential benefits are significant even with an 8-bit quantized representation. On average, skipping all the zero valued neurons would eliminate only 30\% of the terms whereas \PRA would remove up to 71\% of the terms.

In summary, this section corroborated past observations that: a)~many neuron values are zero~\cite{han_eie:_2016,albericio:cnvlutin,isscc_2016_chen_eyeriss, Reagen2016}, and b)~only close to a half of the computations performed traditionally is needed if numerical precision is properly adjusted~\cite{Stripes-CAL}. It further showed that far less computations are really needed, 10\% and 29\% on average for the 16-bit fixed-point and 8-bit quantized representations respectively, if only the essential neuron bits were processed. Finally, software can boost the opportunities for savings by communicating per layer precisions.




\section{\PRA: A Simplified Example}
\label{sec:example}

\begin{figure*}
\centering
\centering
\includegraphics[width=0.7\textwidth]{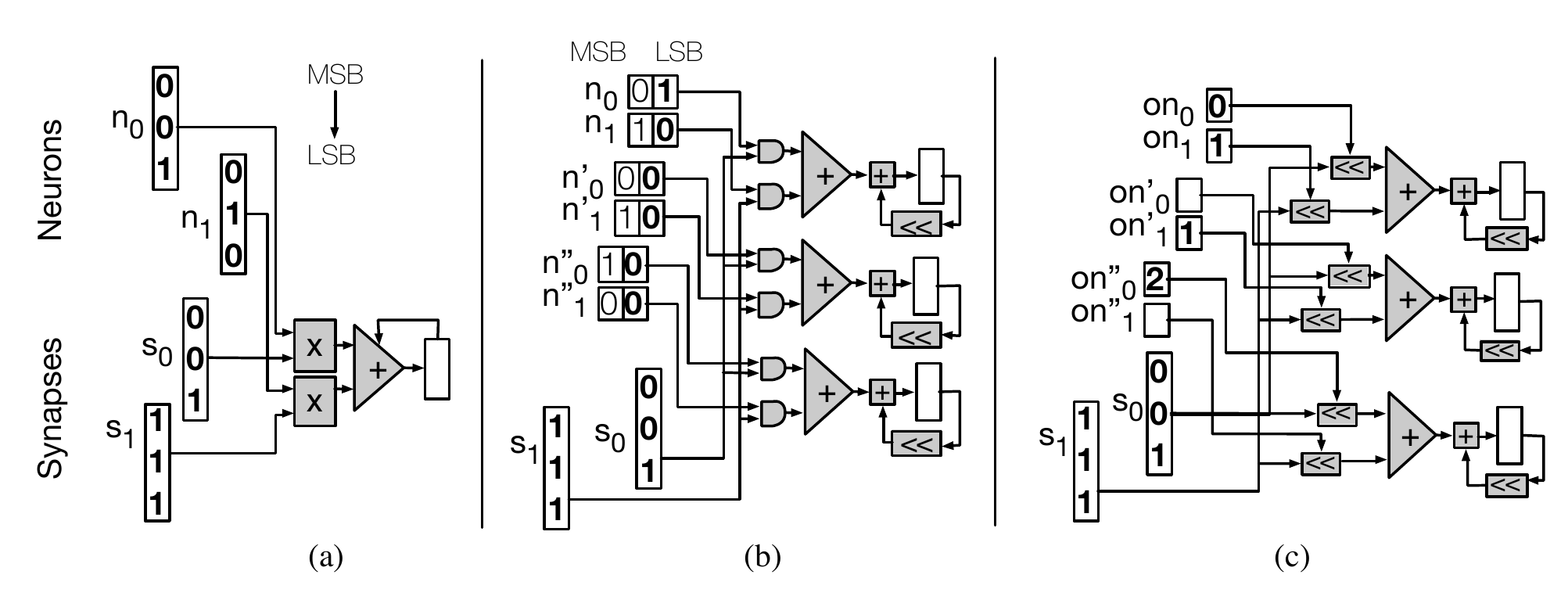}
\caption{ a)~Bit-parallel unit. b) Bit-serial unit with equivalent throughput (\textit{Stripes}\cite{Stripes-CAL}). c)~\PRA unit with equivalent throughput where only essential information is processed. 
\label{fig:tiny-example}
}
\end{figure*}

This section illustrates the idea behind \PRA via a simplified example. For the purposes of this discussion suffices to know that in a convolutional layer there are typically hundreds to thousands of neurons, each multiplied with a corresponding synapse, and that the synapses are reused several times. Section~\ref{sec:convolution} describes the relevant computations in more detail.


The bit-parallel unit of Figure~\ref{fig:tiny-example}a multiplies two neurons with their respective synapses and via an adder reduces the two products. The unit reads \textit{all} neuron and synapse bits,  $(n_0=001_{(2)}, n_1=010_{(2)})$ and $(s_0=001_{(2)}, s_1=111_{(2)})$ respectively in a single cycle. As a result, the two sources of inefficiency EoP and LoE manifest here:  $n_0$ and $n_1$ are represented using 3 bits instead of 2 respectively due to EoP. Even in 2 bits, they each contain a zero bit due to LoE. As a result, four ineffectual terms are processed when using standard multipliers such as those derived from the Shift and Add algorithm. In general, given $N$ neuron and synapse pairs, this unit will take $\lceil{N/2}\rceil$ cycles to process them regardless of their precision and the essential bit content of the neurons.

The hybrid, bit-serial-neuron/bit-parallel-synapse unit in Figure~\ref{fig:tiny-example}b is representative of \STR which tackles EoP. 
Each cycle, the unit processes one bit from each neuron and hence it takes three cycles to compute the convolution when the neurons are represented using 3 bits each, a slowdown of 3x over the bit-parallel engine.
To match the throughput of the bit-parallel engine of  Figure~\ref{fig:tiny-example}a, \STR takes advantage of synapse reuse and processes multiple neurons groups in parallel. In this example, six neurons $(n_0=001_{(2)}, n_1=010_{(2)}, n'_0=000_{(2)}, n'_1=010_{(2)}, n''_0=010_{(2)}, n''_1=000_{(2)})$ are combined with the two synapses as shown. Starting from the least significant position, each cycle one bit per neuron is \textit{ANDed} with the corresponding synapse. The six AND results are added via the reduction tree and the result is accumulated after being shifted by one bit. 
Since the specific neuron values could be represented all using 2 bits, \STR would need 2 cycles to process all six products compared to the 3 cycles needed by the bit-parallel system, a $3/2\times$ speedup.
However, \STRL still processes some ineffectual terms. For example, in the first cycle, 4 of the 6 terms are zero yet they are added via the adder tree, wasting computing resources and energy. 

Figure~\ref{fig:tiny-example}c shows a simplified \PRAS engine. In this example, neurons are no longer represented as vectors of bits but as vectors of offsets of the essential bits. For example, neuron $n_0=001_{(2)}$ is represented as $on_0=(0)$, and a neuron value of $111_{(2)}$ would be represented as $(2,1,0)$. An out-of-band bit (wire) not shown indicates the neuron's end. A shifter per neuron uses the offsets to effectively multiply the corresponding synapse with the respective power of 2 before passing it to the adder tree. As a result, \PRAS processes only the non-zero terms avoiding all ineffectual computations that were due to EoP or LoE.
For this example, \PRAS would process six neuron and synapse pairs in a single cycle, a speedup of $3\times$ over the bit-parallel engine.
\section{Background}
\label{sec:background}
This work presents \PRA as a modification 
of the state-of-the-art \textit{DaDianNao} accelerator. Accordingly, this section provides the necessary background information: Section~\ref{sec:convolution} reviews the operation of convolutional layers, and Section~\ref{sec:baseline} overviews \DaDN 
and how it processes convolutional layers. 

\subsection{Convolutional Layer Computation} 
\label{sec:convolution}
\newcommand{\BBj}[1]{ { \mathbb{#1} } }

A convolutional layer processes and produces neuron arrays, that is 3D arrays of real numbers. The layer applies $N$ 3D filters in a sliding window fashion using a constant stride $S$ to produce an output 3D array. 
The  input array contains $N_x\times N_y \times I$ \textit{neurons}. Each of the $N$ filters, contains $F_x \times F_y \times I$  \textit{synapses} which are also real numbers. The output neuron array dimensions are $O_x \times O_y \times N$, that is its depth equals the filter count. Each filter corresponds to a desired \textit{feature} and the goal of the layer is to determine where in the input neuron array these features appear. Accordingly, each constituent 2D array along the $i$ dimension of the output neuron array corresponds to a \textit{feature}. To calculate an output neuron, the layer applies one filter over a \textit{window}, a filter-sized, or $F_x \times F_y \times I$ sub-array of the input neuron array. If $n(x,y,i)$ and $o(x,y,i)$ are respectively input and output neurons, and $s^f(x,y,i)$ are the synapses of filter ${f}$.  
The output neuron at position $(\BBj{k},\BBj{l},\BBj{f})$ is given by:

\begin{gather*}
\small
\underbrace{o(\BBj{k},\BBj{l},\BBj{f})}_{\substack{output\\ neuron}} = 
\underbrace{
\sum_{y=0}^{F_y-1} 
\sum_{x=0}^{F_x-1} 
\sum_{i=0}^{I-1} 
\underbrace{s^{\BBj{f}}(y,x,i)}_{synapse} \times 
\underbrace{n(y+\BBj{l} \times S, x+\BBj{k} \times S,i)}_{input\ neuron}
}_{window}
\label{eq:convolution}
\end{gather*}

The layer applies filters repeatedly over different windows positioned along the X and Y dimensions using a constant stride $S$, and there is one output neuron per window and filter. Accordingly, the output neuron array dimensions are $O_x=(I_x-F_x)/S+1$, $O_y=(I_y-F_y)/S+1$, and $O_i = N$.   . 

\subsubsection{Terminology -- Bricks and Pallets:}
For clarity, in what follows the term \textit{brick} refers to a set of 16 elements of a 3D neuron or synapse array which are contiguous along the $i$ dimension, e.g., $n(x,y,i)...n(x,y,i+15)$. Bricks will be denoted by their origin element with a $B$ subscript, e.g., $n_B(x,y,i)$. 
The term \textit{pallet} refers to a set of 16 bricks corresponding to adjacent, using a stride $S$, windows along the $x$ or $y$ dimensions, e.g., $n_B(x,y,i)...n_B(x,y+15\times S,i)$ and will be denoted as $n_P(x,y,i)$. The number of neurons per brick, and bricks per pallet are design parameters.

\begin{figure*}
\centering
\subfloat[]{
\centering
\includegraphics[width=0.40\textwidth]{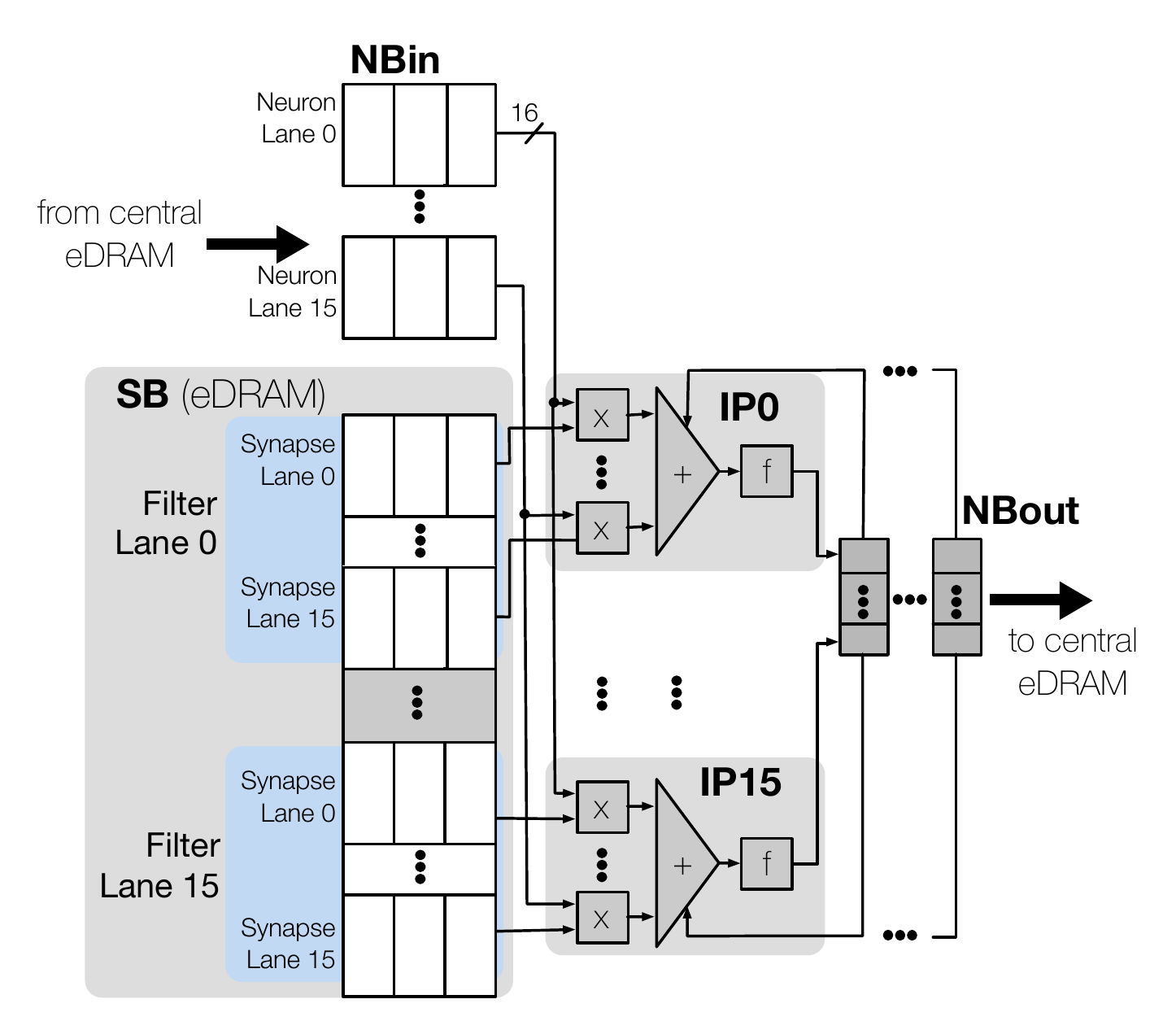}
\label{fig:dadiannao-overview}
}
\subfloat[]{
\centering
\includegraphics[width=0.5\textwidth]{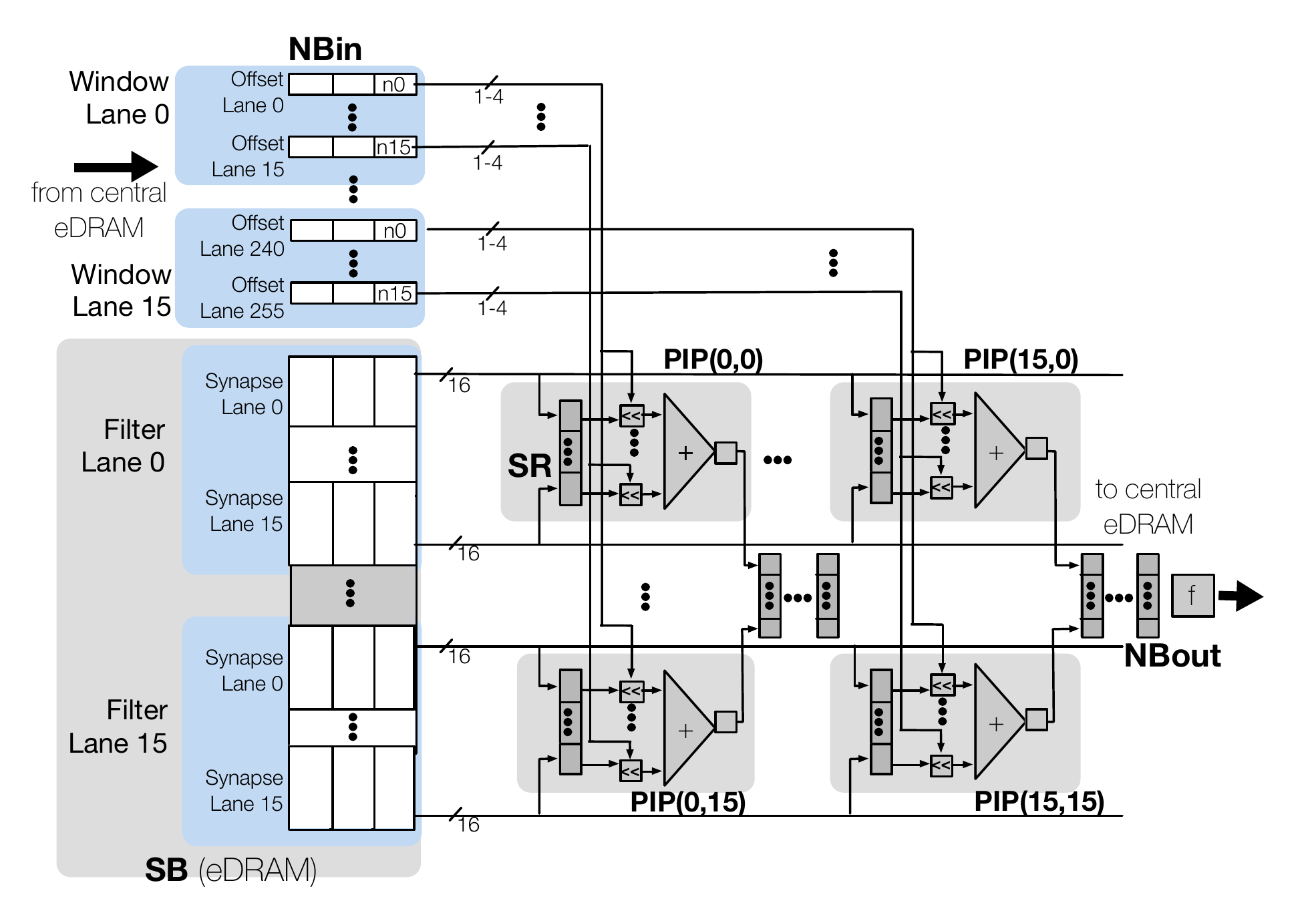}
\label{fig:pragmatic-overview}
}
\caption{a) DaDianNao Tile. b) Pragmatic Tile. 
}
\end{figure*}

%

%
%

\subsection{Baseline System: DaDianNao}
\label{sec:baseline}

\newcommand{\tile}{tile\xspace}
\newcommand{\tiles}{tiles\xspace}
\newcommand{\nfu}{unit\xspace}
\newcommand{\nfus}{units\xspace}
\PRA is demonstrated as a modification of the \textit{DaDianNao} accelerator (\BASE)  proposed by Chen \textit{et al.}~\cite{DaDiannao}. Figure~\ref{fig:dadiannao-overview} shows a \BASE \tile which processes 16 filters concurrently calculating 16 neuron and synapse products per filter for a total of 256 products per cycle. To do, each cycle the \tile accepts 16 synapses per filter for total of 256 synapses, and 16 input neurons. The \tile multiplies each synapse with only one neuron whereas each neuron is multiplied with 16 synapses, one per filter.   The \tile reduces the 16 products into a single partial output neuron per filter, for a total of 16 partial output neurons for the \tile. Each \BASE chip comprises 16 such \tiles, each processing a different set of 16 filters per cycle. Accordingly, each cycle, the whole chip processes 16 neurons and $256\times 16=4K$ synapses producing $16\times 16=256$ partial output neurons.

Internally, each \tile has: 1)~a synapse buffer (SB) that provides 256 synapses per cycle one per synapse lane, 2)~an input neuron buffer (NBin) which provides 16 neurons per cycle through 16 neuron lanes, and 3)~a neuron output buffer (NBout) which accepts 16 partial output neurons per cycle. In the \tile's datapath, or the \textit{Neural Functional Unit} (NFU)  each neuron lane is paired with 16 synapse lanes one from each filter. 
Each  synapse and neuron lane pair feed a multiplier and an adder tree per filter lane reduces the 16 per filter products into a partial sum. 
In all, the filter lanes produce each a partial sum per cycle, for a total of 16 partial output neurons per NFU. 
Once a full window is processed, the 16 resulting sums,  are fed through a non-linear activation function, $f$, to produce the 16 final output neurons.  
The multiplications and reductions needed per cycle are implemented via 256 multipliers one per synapse lane and sixteen 17-input (16 products plus the partial sum from NBout) adder trees one per filter lane. 


\BASE's main goal was minimizing off-chip bandwidth while maximizing on-chip compute utilization. To avoid fetching synapses from off-chip, \BASE uses a 2MB eDRAM SB per \tile for a total of 32MB eDRAM. All inter-layer neuron outputs except for the initial input and the final output are stored in a 4MB shared central eDRAM \textit{Neuron Memory} (NM) which is connected via a broadcast interconnect to the 16 NBin buffers. Off-chip accesses are needed only for reading the input image, the synapses once per layer, and for writing the final output.

Processing starts by reading from external memory the first layer's filter synapses, and the input image. The synapses are distributed over the SBs and the input is stored into NM. Each cycle an input neuron brick is broadcast to all units. Each units reads 16 synapse bricks from its SB and produces a partial output neuron brick which it stores in its NBout. Once computed, the output neurons are stored through NBout to NM and then fed back through the NBins when processing the next layer. Loading the next set of synapses from external memory can be overlapped with the processing of the current layer as necessary.  


\section{\PRA}
\label{sec:bit-prag-computing}


This section presents the \PRA architecture. Section~\ref{sec:prag-approach} describes  \PRAS's processing approach while Section~\ref{sec:prag-tile} describes its organization. Sections~\ref{sec:prag-2stage} and~\ref{sec:prag-synchro} present two optimizations that respectively improve area and performance. For simplicity, the description assumes specific values for various design parameters so that \PRAS performance matches that of the \BASE configuration of Section~\ref{sec:baseline} in the worst case.

\subsection{Approach}
\label{sec:prag-approach}
\PRAS's goal is to process only the essential bits of the input neurons. To do so \PRAS a)~converts, on-the-fly, the input neuron representation into one that contains only the essential bits, and b)~processes one essential bit per neuron and a full 16-bit synapse per cycle.  Since \PRAS processes neuron bits serially, it may take up to 16 cycles to produce a product of a neuron and a synapse. To always match or exceed the performance of the bit-parallel units of \BASE, \PRAS processes more neurons concurrently exploiting the abundant parallelism of the convolutional layers. The remaining of this section describes in turn: 1)~an appropriate neuron representation, 2)~the way \PRAS calculates terms, 3)~how multiple terms are processed concurrently to maintain performance on par with \BASE in the worst case, and 4)~how \PRAS's units are supplied with the necessary neurons from NM. 

\subsubsection{Input Neuron Representation}
\PRAS starts with an input neuron representation where it is straightforward to identify the next essential bit each cycle. One such representation is an explicit list of \textit{oneffsets}, that is of the constituent powers of two. For example, a neuron  $n=5.5_{(10)}=0101.1_{(2)}$ would be represented as $n=(2,0,-1)$. In the implementation described herein, neurons are stored in 16-bit fixed-point in NM, and converted on-the-fly in the \PRAS representation as they are broadcast to the tiles. A single oneffset is processed per neuron per cycle. Each oneffset is represented as $(pow,eon)$ where $pow$ is a 4-bit value and $eon$ a single bit which if set indicates the end of a neuron. For example, $n=101_{(2)}$ is represented as $n^{\PRAS}=((0010,0)(0000,1))$.
In the worst case, all bits of an input neuron would be 1 and hence its \PRAS representation would contain 16 oneffsets. 

\subsubsection{Calculating a Term}
\PRAS calculates the product of synapse $s$ and neuron $n$ as:
\begin{equation*}
s\times n= \sum\limits_{\forall f \in n^\PRAS}{s\times 2^f}= \sum_{\forall f\in n^\PRAS}{(n \ll f)}
\end{equation*}

That is, each cycle, the synapse $s$ multiplied by $f$, the next constituent power two of $n$, and the result is accumulated. This multiplication can be implemented as a shift and an AND.

\subsubsection{Boosting Compute Bandwidth over \BASE}To match \BASE's performance  \PRAS needs to process the same number of effectual terms per cycle.
Each \BASE \tile calculates 256 neuron and synapse products per cycle, or $256\times 16=4K$ terms. While  most of these terms will be in practice ineffectual, to guarantee that \PRAS always performs as well as \BASE it should process $4K$ terms per cycle. For the time being let us assume that all neurons contain the same number of essential bits, so that when processing multiple neurons in parallel, all units complete at the same time and thus can proceed with the next set of neurons in sync. The next section will relax this constraint.

Since \PRAS processes neurons bits serially, it produces one term per neuron bit and synapse pair and thus needs to process $4K$ such pairs concurrently.  The choice of which $4K$ neuron bit and synapse pairs to process concurrently can adversely affect complexity and performance. For example, it could force an increase in SB capacity and width, or an increase in NM width, or be ineffective due to unit underutilization given the commonly used layer sizes.

Fortunately, it is possible to avoid increasing the capacity and the width of the SB and the NM while keeping the units utilized as in \BASE. Specifically, a \PRAS tile can read 16 synapse bricks and the equivalent of 256 neuron bits as \BASE's tiles do (\BASE processes 16 16-bit neurons or 256 neuron bits per cycle). Specifically, as in \BASE, each \PRAS tile processes 16 synapse bricks concurrently, one per filter. However, differently than \BASE where the 16 synapse bricks are combined with just one neuron brick which is processed bit-parallel, \PRAS combines each synapse brick with 16 neuron bricks, one from each of 16 windows, which are processed bit-serially. The same 16 neuron bricks are combined with all synapse bricks. These neuron bricks form a \textit{pallet} enabling the same synapse brick to be combined with all. 
For example, in a single cycle a \PRAS title processing filters $0$ through $15$ could combine combine $s_B^0(x,y,0), ..., s_B^15(x,y,0)$ with $n_B^\PRAS(x,y,0), n_B^\PRAS(x+2,y,0), ... n_B^\PRAS(x+31,y,0)$ assuming a layer with a stride of 2. In this case, $s^4(x,y,2)$ would be paired with $n^\PRAS(x,y,2)$, $n^\PRAS(x+2,y,2)$, ...,  $n^\PRAS(x+31,y,2)$ to produce the output neurons $on(x,y,4)$ through $on(x+15,y,4)$.

As the example illustrates, this approach allows each synapse to be combined with one neuron per window whereas in \BASE each synapse is combined with one neuron only. In total, 256 essential neuron bits are processed per cycle and given that there are 256 synapses and 16 windows, \PRAS processes $256\times 16=4K$ neuron bit and synapse pairs, or terms per cycle producing 256 partial output neurons, 16 per filter, or 16 partial output neuron bricks per cycle.

\subsubsection{Supplying the Input Neuron and Synapse Bricks}
\label{sec:datasupply}
Thus far it was assumed that all input neurons have the same number of essential bits. Under this assumption, all neuron lanes complete processing their terms at the same time, allowing \PRAS to move on to the next neuron pallet and the next set of synapse bricks in one step. This allows \PRAS to reuse \STR's approach for fetching the next pallet from the single-ported NM~\cite{Stripes-CAL}. Briefly, with unit stride the 256 neurons would be typically all stored in the same NM row or at most over two adjacent NM rows and thus can be fetched in at most two cycles. When the stride is more than one, the neurons will be spread over multiple rows and thus multiple cycles will be needed to fetch them all. Fortunately, fetching the next pallet can be overlapped with processing the current one. Accordingly, if it takes $NM_C$ to access the next pallet from NM, while the current pallet requires $P_C$ cycles to process, the next pallet will begin processing after $max(NM_C, P_C)$ cycles. When $NM_C > P_C$ performance is lost waiting for NM.

In practice it highly unlikely that all neurons will have the same number of essential bits. In general, each neuron lane if left unrestricted will advance at a different rate. In the worst case, each neuron lane may end up needing neurons from a different neuron brick, thus breaking \PRAS's ability to reuse the same synapse brick.  This is undesirable if not impractical as it would require partitioning and replicating the SB so that 4K unrelated synapses could be read per cycle, and it would also increase NM complexity and bandwidth.  

Fortunately, these complexities can be avoided with \textit{pallet-level neuron lane synchronization} where all neuron lanes ``wait'' (a neuron lane that has detected the end of its neuron forces zero terms while waiting) for the one with the most essential bits to finish before proceeding with the next pallet. Under this approach, it does not matter which bits are essential per neuron, only how many exist.
Since, it is unlikely that most pallets will contain a neuron with 16 essential terms, \PRAS will improve performance  over \BASE. Section~\ref{sec:eval-2-stage} will show that in practice, this approach improves performance over \BASE and \STR. Section~\ref{sec:prag-synchro} will discuss finer-grain synchronization schemes that lead to even better performance. Before doing so, however, the intervening sections detail \PRAS's design.



\begin{figure}
\centering
\includegraphics[width=\linewidth]{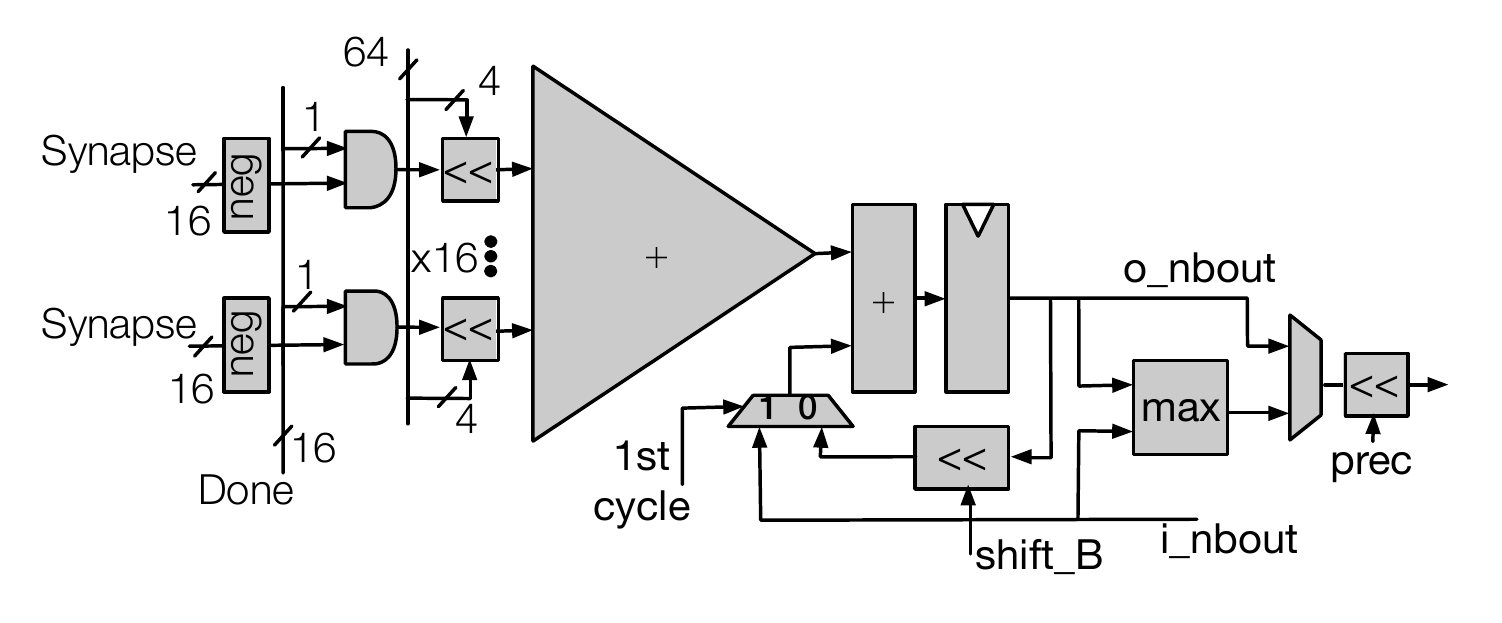}
\caption{Pragmatic Inner Product Unit.}
\label{fig:PIP-internals}
\end{figure}

\begin{figure*}
\centering
\includegraphics[width=0.9\linewidth]{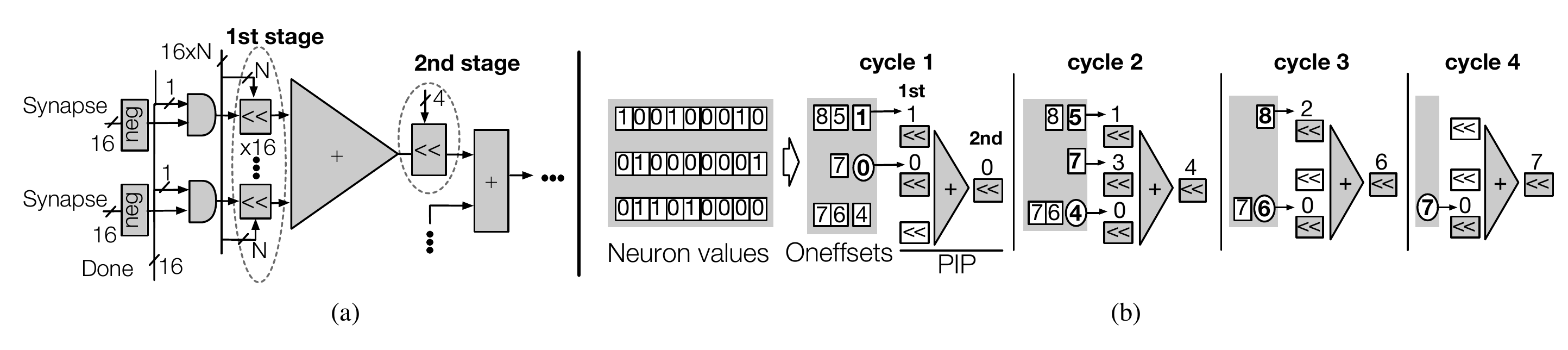}
\caption{2-stage shifting. a) \textmd{Modified PIP.} b) \textmd{ Example: Processing three 9-bit synapse and neuron pairs with $L=2$. The oneffset generator reads the neuron values, and produces a set of three oneffests per cycle. 
Each cycle, the control logic, which is shared and amortized across the entire column of PIPs, compares the oneffsets being processed, (1, 0, 4) in the first cycle of our example and picks the lowest, 0, indicated by a circle. This minimum oneffset controls the second stage shifter. The control subtracts this offset from all three oneffsets. The difference per oneffset, as long as it is less than $2^L$ controls the corresponding first level shifter. In the first cycle, the two shifters at the top are fed with values $1-0=1$ and $0-0=0$, while the shifter at the bottom is stalled given that it is not able to handle a shift by $4-0=4$. On cycle 2, the oneffsets are (6,7,4) and 4 is now the minimum, which controls the 2nd stage shifter, while (1, 3, 0) control the first-level shifters. On cycle 3, only the first and the third neurons still have oneffsets to process. The computation  finishes in cycle 4 when the last oneffset of the third neuron controls the shifters. } }
\label{fig:2-stage-example}
\end{figure*}

\subsection{Tile Organization and Operation}
\label{sec:prag-tile}
Figure~\ref{fig:pragmatic-overview} shows the \PRA tile architecture which comprises an array of $16\times 16 = 256$ \textit{pragmatic inner product units (PIPs)}. PIP(i,j) 
processes a neuron oneffset from the i-th window and its corresponding synapse from the j-th filter. Specifically, all the PIPs along the i-th row receive the same synapse brick belonging to the i-th filter and all PIPs along the j-th column receive an oneffset from each neuron from one neuron brick belonging to the j-th window. 

The necessary neuron oneffsets are read from NBin where they have been placed by the Dispatcher and the Oneffset generators units as Section~\ref{sec:oneffsetgen} explains. Every cycle NBin sends 256 oneffsets 16 per window lane. All the PIPs in a column receive the same 16 oneffsets, corresponding to the neurons of a single window. When the tile starts to process a new neuron pallet, 256 synapses are read from SB through its 256 synapse lanes as in \BASE and are stored in the synapse registers (SR) of each PIP. 
The synapse and oneffsets are then processed by the PIPs as the next section describes.



\subsubsection{Pragmatic Inner-Product Unit}
\label{sec:prag-unit}
Figure~\ref{fig:PIP-internals} shows the PIP internals. 
Every cycle, 16 synapses are combined with their corresponding oneffsets. Each oneffsets controls a shifter effectively multiplying the synapse with a power of two. The shifted synapses are reduced via the adder tree. An AND gate per synapse supports the injection of a null terms when necessary.
In the most straightforward design, the oneffsets use 4-bits, each shifter accepts a 16-bit synapse and can shift it by up to 15 bit positions producing a 31-bit output. Finally, the adder tree accepts 31-bit inputs. Section~\ref{sec:prag-2stage} presents an enhanced design that requires narrower components improving area and energy.


\begin{figure*}
\centering
\includegraphics[width=0.9\linewidth]{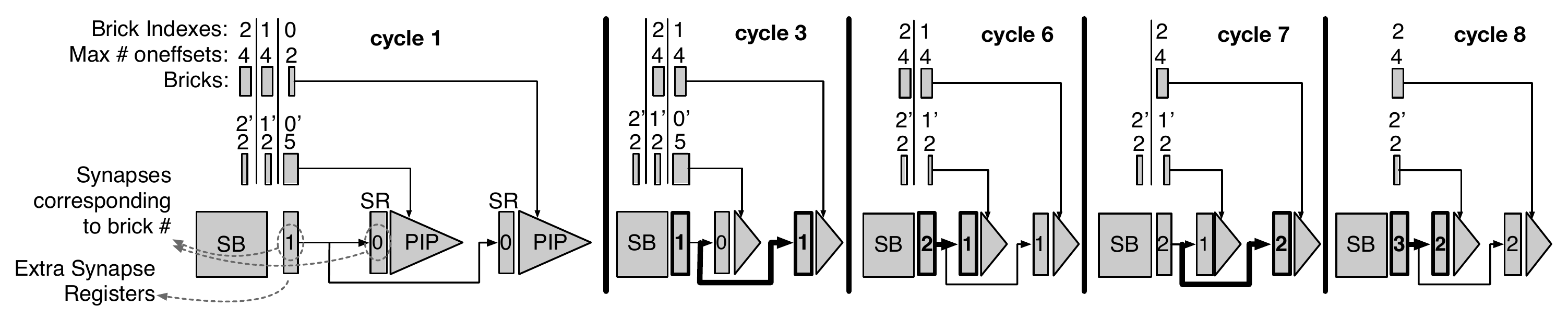}
\caption{Per-column synchronization example: \textmd{one extra synapse register and 1x2 PIP array capable of processing two windows in parallel. The two numbers per brick show: the first from the top is the brick's index, $(0,1,2)$ and  $(0',1',2')$ for the bricks of the first and second window. The second is the maximum count of oneffsets in its neurons, $(2,4,4)$ and $(5,2,2)$ respectively. The numbers in 
the registers indicate the index of the corresponding bricks, i.e., a synapse register containing a $K$ stores the synapses corresponding to neurons bricks with indexes $K$ and $K'$. In cycles 3 to 8, thicker lines indicate registers being loaded or wires being used. }}
\label{fig:synchro-example}
\end{figure*}

\subsection{Dispatcher and  Oneffset Generators} 
\label{sec:oneffsetgen} 
The \textit{Dispatcher} reads 16 neuron bricks from NM, as expected by the \PRAS tiles. The  \textit{oneffset generator}   converts their neurons on-the-fly to the oneffset representation, and broadcasts one oneffset per neuron per cycle for a total of 256 oneffsets to all titles. Fetching and assembling the 16 neuron bricks from NM is akin to fetching words with a stride of $S$ from a cache structure. As Section~\ref{sec:datasupply} discussed this can take multiple cycles depending on the stride and alignment of the initial neuron brick. \PRAS uses the same dispatcher design as \STR~\cite{Stripes-CAL}.

Once the 16 neuron bricks have been collected, 256 oneffset generators operate in parallel to locate and communicate the next oneffset per neuron. A straightforward 16-bit leading one detector is sufficient. The latency of the oneffset generators and the dispatcher can be readily hidden as they can be pipelined as desired overlapping them with processing in the \PRAS tiles.

\subsection{2-Stage Shifting}
\label{sec:prag-2stage}
Any shift can be performed in two stages as two smaller shifts: $a\ll K = a \ll (K'+C) = ((a \ll K') \ll C)$. 
Thus, to shift and add $T$ synapses by different offsets $K_0,..., K_T$, we can decompose the offsets into sums with a common term $C$, e.g., $K_i=K'_i + C$. Accordingly, PIP processing can be rearranged using a two stage processing where the first stage uses the synapse specific offsets $K'_i$, and the second stage, the common across all synapses offset $C$:

\begin{gather*}
\small
\underbrace{\sum^T_{i}{(S_i\ll K_i)}}_{1-stage\ shifting}=\sum^T_{i}{(S_i\ll (K'_i+C))}=\\
(\underbrace{\sum^T_{i}{(S_i\ll K'_i)}}_{1st\ stage})\underbrace{\ll C}_{2nd\ stage} 
\end{gather*}


This arrangement can be used to reduce the width of the synapse shifters and of the adder tree by sharing one common shifter after the adder tree as Figure~\ref{fig:2-stage-example}a shows. A design parameter, $L$, defines the number of bits controlling the synapse shifters. Meaning the design can process oneffsets which differ by less than $2^L$ in a single cycle. This reduces the size of the synapse shifters and reduces the size of the adder tree to support terms of $16+2^L-1$ bits only. As Section~\ref{sec:eval-2-stage}  shows, this design reduces the area of the shifters and the adder trees which are the largest components of the PIP. 
Figure~\ref{fig:2-stage-example}b shows an example illustrating how this PIP can handle any combination of oneffsets. Section~\ref{sec:eval-2-stage} studies the impact of $L$ on cost and performance.


\subsection{Per-Column Neuron Lane Synchronization}
\label{sec:prag-synchro}
The pallet neuron lane synchronization scheme of Section~\ref{sec:datasupply} is one of many possible synchronization schemes. 
Finer-grain neuron lane synchronization schemes are possible leading to higher performance albeit at a cost. This section presents \textit{per column} neuron lane synchronization, an appealing scheme that, as Section~\ref{sec:eval-synchro} shows, enhances performance at little additional cost.


Here each PIP column operates independently but all the PIPs along the same column wait for the neuron with the most essential bits before proceeding to the next neuron brick. Since the PIPs along the same column operate in sync, they all process one set of 16 synapse bricks which can be read using the existing SB interface.
However, given that different PIP  columns operate now out-of-sync, the SB would be read a higher number of times and become a bottleneck.

There are two concerns: 1)~different PIP columns may need to perform two independent SB reads while there are only one SB port and one common bus connecting the PIP array to the SB, and 2)~there will be repeat accesses to SB that will increase SB energy, while the SB is already a major contribution of energy consumption. These concerns are addressed as follows: 1)~only one SB access can proceed per cycle thus a PIP column may need to wait when collisions occur. This way, we do not need an extra SB read port nor an extra set of 4K wires from the SB to the PIP array. 2)~A set of SRAM registers, or \textit{synapse set registers} (SSRs) are introduced in front of the SB each holding a recently read set of 16 synapse bricks. Since all PIP columns will eventually need the same set of synapse bricks, temporarily buffering them avoids fetching them repeatedly from the SB. Once a synapse set has been read into an SSR, it stays there until all PIP columns have copied it (a 4-bit down counter is sufficient for tracking how many PIP columns have yet to read the synapse set). This policy guarantees that the SB is accessed the same number of times as in \BASE. However, stalls may incur as a PIP column has to be able to store a new set of synapses into an SSR when it reads it from  the SB. Figure~\ref{fig:synchro-example} shows an example. Section~\ref{sec:eval-synchro} evaluates this design. 

Since each neuron lane advances independently, in the worst case, the dispatcher may need to fetch 16 independent neuron bricks each from a different pallet. The Dispatcher can buffer those pallets to avoid rereading NM, which would, at worst,  require a 256 pallet buffer. However, given that the number SSRs restricts how far apart the PIP columns can be, and since Section~\ref{sec:eval-synchro} shows that only one SSR is sufficient, a two pallet buffer in the dispatcher is all that is needed.




\subsection{The Role of Software}
\label{sec:software}
\PRAS enables  an additional dimension upon which hardware and software can attempt to further boost performance and energy efficiency, that of controlling the essential neuron value content. 
This work investigates a software guided approach where the precision requirements of each layer are used to zero out a number of prefix and suffix bits at the output of each layer. Using the profiling method of Judd \textit{et al.,}~\cite{judd:reduced}, software communicates the precisions needed by each layer as meta-data. The hardware trims the output neurons before writing them to NM using AND gates and precision derived bit masks. 
\section{Evaluation}
\label{sec:evaluation}
The performance, area and energy efficiency of \PRA is compared against \BASE~\cite{DaDiannao}  and \STRL~\cite{Stripes-CAL}, two state-of-the-art DNN accelerators. \BASE is the fastest bit-parallel accelerator proposed to date that processes all neuron regardless of theirs values, and \STR improves upon \BASE by exploiting the per layer precision requirements of DNNs. \textit{Cnvlutin}  improves upon \BASE by skipping most zero-valued neurons~\cite{albericio:cnvlutin}, however, \STRL has been shown to outperform it. 

The rest of this section is organized as follows: Section~\ref{sec:methodology} presents the the experimental methodology. Sections~\ref{sec:eval-2-stage} and ~\ref{sec:eval-synchro} explore the \PRAS design space considering respectively single- and 2-stage  shifting configurations, and column synchronization. Section~\ref{sec:energy}  reports energy efficiency for the best configuration. Section~\ref{sec:evalsoft} analyzes the contribution of the software provided precisions. Finally, Section~\ref{sec:quantized} reports performance for designs using an 8-bit quantized representation.


\subsection{Methodology}
\label{sec:methodology}






All systems were modelled using the same methodology for consistency. A custom cycle-accurate simulator models execution time. Computation was scheduled such that all designs see the same reuse of synapses and thus the same SB read energy.
To estimate power and area, all designs were synthesized with the Synopsis Design Compiler~\cite{synopsys_site} for a TSMC 65nm library. The NBin and NBout SRAM buffers were modelled using CACTI~\cite{Muralimanohar_cacti6.0:}. The eDRAM area and energy were modelled with \textit{Destiny}~\cite{destiny}. To compare against \STR, the per layer numerical representation requirements reported in Table~\ref{tab:best_mixed} were found 
using the methodology of Judd et al.~\cite{Stripes-CAL}. All \PRAS configurations studied exploit software provided precisions as per Section~\ref{sec:software}.  Section~\ref{sec:evalsoft} analyzes the impact of this information on overall performance. All performance measurements are for the convolutional layers only which account for more than 92\% of the overall execution time in \BASE~\cite{DaDiannao}.
\PRAS does not affect the execution time of the remaining layers.

\begin{table}[!t]
\small
    \centering
    \begin{tabular}{|l|p{4.5cm}|}
    \hline
   &   \textbf{Per Layer}  \\
\textbf{Network} &  \textbf{Neuron Precision in Bits}  \\ \hline
        \hline  

AlexNet         &   9-8-5-5-7                                           \\ \hline
NiN             &  8-8-8-9-7-8-8-9-9-8-8-8                             \\ \hline
GoogLeNet       &  10-8-10-9-8-10-9-8-9-10-7                           \\ \hline
VGG\_M          &  7-7-7-8-7                                           \\ \hline
VGG\_S          &  7-8-9-7-9                                           \\ \hline
VGG\_19         &  12-12-12-11-12-10-11-11-13-12-13-13-13-13-13-13     \\ \hline
\end{tabular}
\caption{
Per layer neuron precision profiles. 
}
\label{tab:best_mixed}
\end{table}



\subsection{Single- and 2-stage Shifting}
\label{sec:eval-2-stage}
This section evaluates the single-stage shifting \PRAS configuration of Sections~\ref{sec:prag-approach}--~\ref{sec:prag-tile} , and the 2-stage shifting variants of Section~\ref{sec:prag-2stage}. Section~\ref{sec:2sperf} reports performance while Section~\ref{sec:2stageenergy} reports area and power. In this section, All \PRAS systems  use pallet synchronization.

\subsubsection{Performance:}
\label{sec:2sperf}
\begin{figure}[t]
\centering
\includegraphics[width=0.45\textwidth]{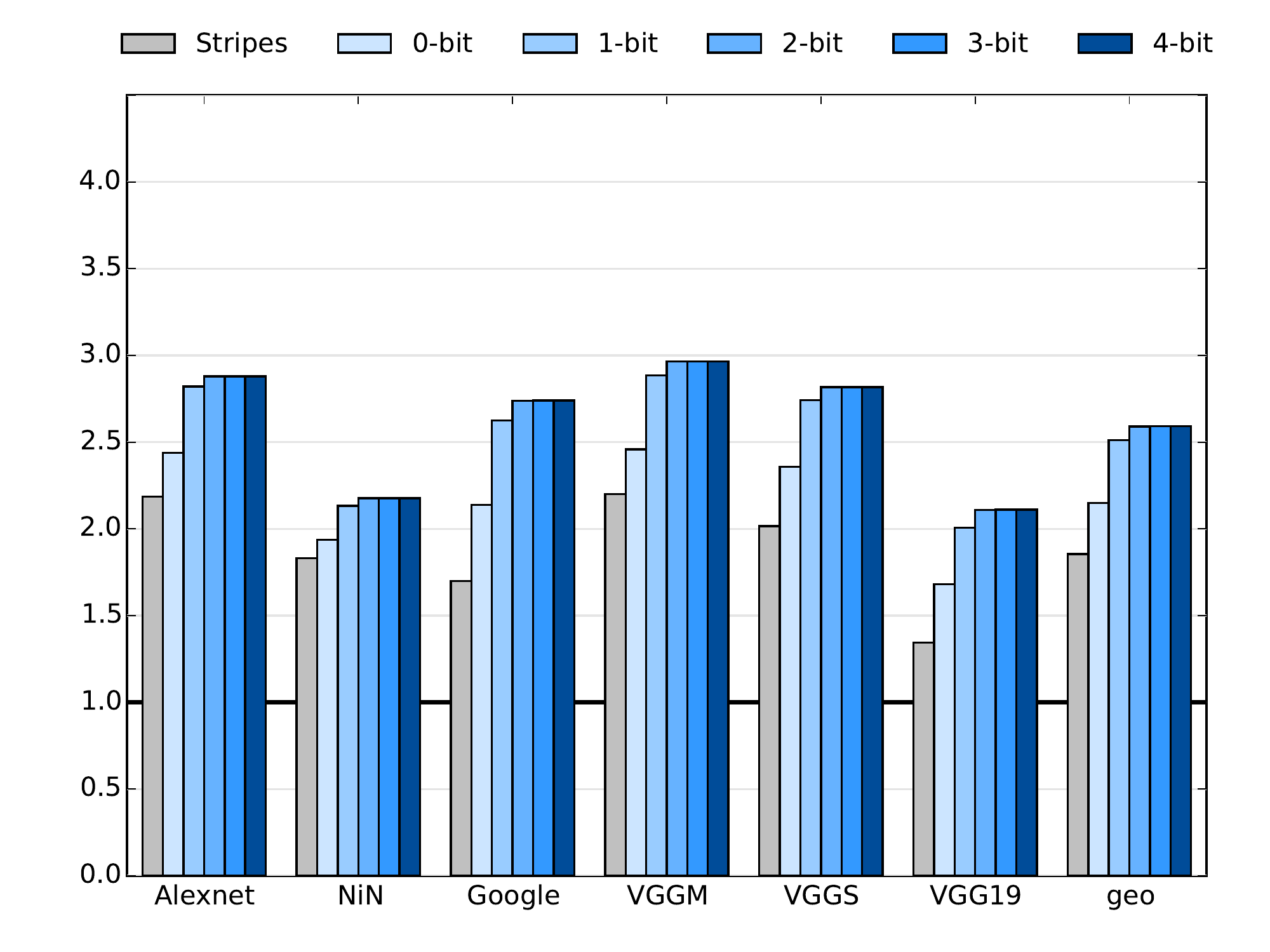}
\caption{\PRA's performance relative to DaDianNao using 2-stage shifting and per-pallet synchronization. 
}
\label{fig:2-stage-speedup}
\end{figure}

Figure~\ref{fig:2-stage-speedup} shows the  performance of \STR (leftmost bars) and of \PRAS variants relative to \BASE. The \PRAS systems are labelled with the number of bits used to operate the first-stage, synapse shifters, e.g., the synapse shifters of \textit{``2-bit''}  , or $\PRAS_{2b}$, are able to shift to four bit positions (0--3). ``4-bit''  or $\PRAS_{4b}$, is the single-stage \PRA, or $\PRAS_{single}$ of Sections~\ref{sec:prag-approach}--~\ref{sec:prag-tile} whose synapse shifters can shift to 16 bit positions (0--15). It has no second stage shifter.  


$\PRAS_{single}$  improves performance  by $2.59\times$ on average over \BASE compared to the $1.85\times$ average improvement with \STR. Performance improvements  over \BASE vary from $2.11\times$ for VGG19 to $2.97\times$ for VGGM.
As expected the 2-stage \PRAS variants offer slightly lower performance than $\PRAS_{single}$, however, performance with $\PRAS_{2b}$ and $\PRAS_{3b}$ is always within 0.2\% of $\PRAS_{single}$. Even $\PRAS_{0b}$  which does not include any synapse shifters outperforms \STR by 20\% on average. Given a set of oneffsets,  $\PRAS_{0b}$ will accommodate the minimum non-zero oneffset per cycle via its second level shifter. 

\subsubsection{Area and Power:}
\label{sec:2stageenergy}
Table~\ref{tab:areapower-2-stage} shows the absolute and relative to \BASE area and power. Two area measurements are reported: 1)~for the unit excluding the SB, NBin and NBout memory blocks, and 2)~for the whole chip comprising 16 units and all memory blocks. Since SB\ and NM dominate chip area, the per area area overheads  Given the performance advantage of \PRAS, the area and power overheads are justified. $\PRAS_{2b}$ is particularly appealing as its overall area cost over $BASE$ is only $1.35\times$ and its power $2.03\times$ while its performance is  $2.59\times$ on average. Accordingly, we restrict attention to this configuration in the rest of this evaluation. 

\begin{table}[t]
\centering
\footnotesize
\begin{tabular}{|c||c|c|c|c|c|c|c|}
\hline
\centering
& DDN & STR & 0-bit & 1-bit & 2-bit & 3-bit & 4-bit \\
\hline
\hline
Area U. & 1.55 & 3.05 & 3.11 & 3.16 & 3.54 & 4.41 & 5.75 \\
$\Delta$ Area U.  & 1.00 & 1.97 & 2.01 & 2.04 & 2.29 & 2.85 & 3.71 \\
\hline
Area T. & 90 & 114 & 115 & 116 & 122 & 136 & 157 \\
$\Delta$ Area T.  & 1.00 & 1.27 & 1.28 & 1.29 & 1.35 & 1.51 & 1.75 \\
\hline
\hline
Power T. & 18.8 & 30.2 & 31.4 & 34.5 & 38.2 & 43.8 & 51.6 \\
$\Delta$ Power T.  & 1.00 & 1.60 & 1.67 & 1.83 & 2.03 & 2.33 & 2.74 \\
\hline
\end{tabular}
\caption{Area [$mm^2$] and power [$W$] for the  unit and the whole chip. Pallet synchronization. }
\label{tab:areapower-2-stage}
\end{table}

\begin{figure}[t]
\centering
\includegraphics[width=0.45\textwidth]{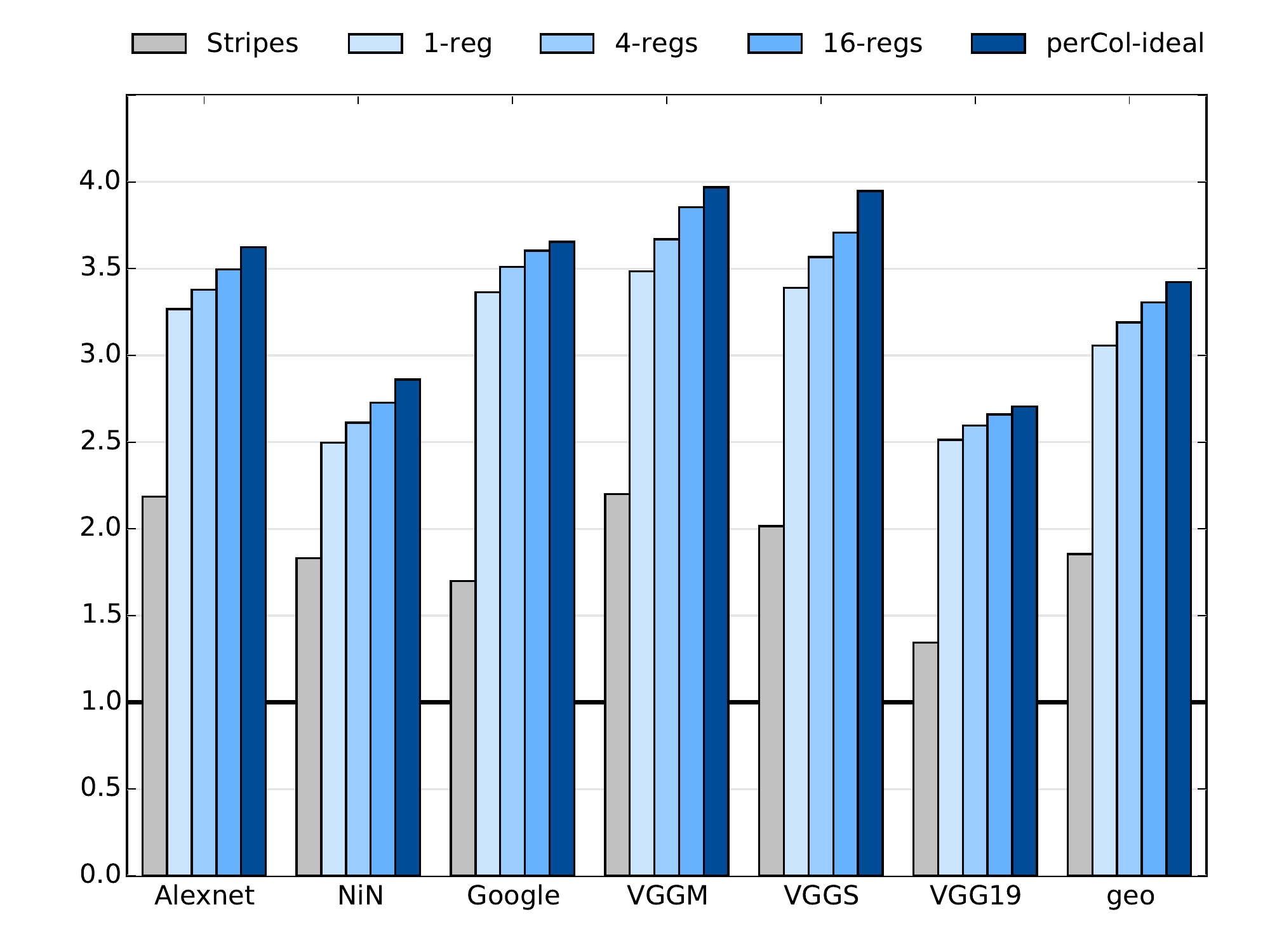}
\caption{Relative performance of $\PRAS_{2b}$ with column synchronization and as a function of the SB registers used. 
}
\label{fig:ESR-speedup}
\end{figure}

\begin{table}[t]
\centering
\footnotesize
\begin{tabular}{|c||c|c|c|c|c|}
\hline
\centering
& DDN & STR & 1-reg & 4-reg & 16-reg \\
\hline
\hline
Area U. & 1.55 & 3.05 & 3.58 & 3.73 & 4.33 \\
$\Delta$ Area U.  & 1.00 & 1.97 & 2.31 & 2.41 & 2.79 \\
\hline
Area T. & 90 & 114 & 122 & 125 & 134 \\
$\Delta$ Area T.  & 1.00 & 1.27 & 1.36 & 1.39 & 1.49 \\
\hline
\hline
Power T. & 18.8 & 30.2 & 38.8 & 40.8 & 49.1 \\
$\Delta$ Power T.  & 1.00 & 1.60 & 2.06 & 2.17 & 2.61 \\
\hline
\end{tabular}
\caption{Area [$mm^2$] and power [$W$] for the  unit and the whole chip for column synchronization and $\PRAS_{2b}$.}
\label{tab:areapower-regs}
\end{table}

\subsection{Per-column synchronization}
\label{sec:eval-synchro}

\subsubsection{Performance:}
Figure~\ref{fig:ESR-speedup} reports the  performance for $\PRAS_{2b}$ with column synchronization and as a function of the number of SSRs as per Section~\ref{sec:prag-synchro}.  of \STRL (first bar of each group) and \PRA (rest of the bars) relative to \BASE. Configuration $\PRAS_{2b}^{xR}$ refers to a configuration using $x$ SSRs. Even $\PRAS_{2b}^{1R}$ boosts performance to $3.1\times$ on average close to the $3.45\times$ that is ideally possible  with $\PRAS_{2b}^{\infty R}$. 

\subsubsection{Area and Power:}
Table~\ref{tab:areapower-regs} reports the area per unit, and the area and power per chip. $\PRAS_{2b}^{1R}$ that offers most performance benefits increases chip area by only $1.35\times$ and power by only $2.19\times$ over \BASE.

\subsection{Energy Efficiency}
\label{sec:energy}
\begin{figure}[t]
\centering
\includegraphics[width=0.45\textwidth]{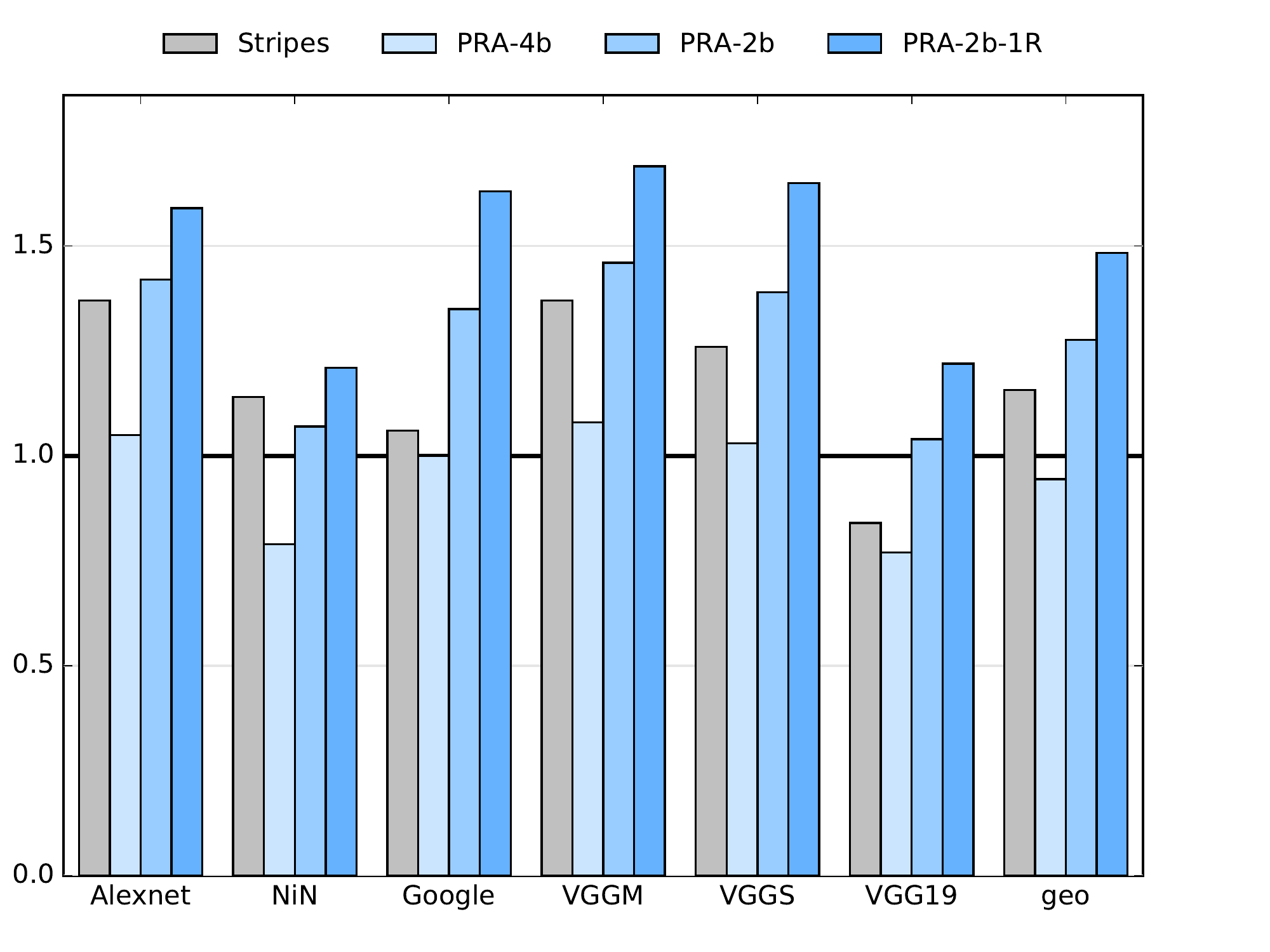}
\caption{Relative energy efficiency}
\label{fig:efficiency}
\end{figure}

Figure~\ref{fig:efficiency} shows the energy efficiency of various configurations of \PRA.
\textit{Energy Efficiency}, or simply \textit{efficiency} for a system \textsc{new} relative to \textsc{base} is defined as the ratio 
$E_{\textsc{base}}/E_{\textsc{new}}$ of the energy required by \textsc{base} to compute all of the convolution layers over that of \textsc{new}.
For the selected networks, \STR is 16\% more efficient than \BASE.  
The power overhead of $\PRAS_{single}$ ($PRA_{4b}$) is more than the speedup resulting in a circuit that is 5\% less efficient than \BASE. 
$\PRAS_{2b}$ reduces that power overhead while maintaining performance yielding an efficiency of 28\%. 
$\PRAS_{2b}^{1R}$ yields the best efficiency at 48\% over \BASE.

\begin{table}[]
\centering
\footnotesize
\begin{tabular}{|r|r|r|r|r|r|r|}
\hline
 \textbf{Alexnet} & \textbf{NiN}   & \textbf{Google} & \textbf{VGGM}  & \textbf{VGGS}  & \textbf{VGG19} & \textbf{AVG}\\ 
\hline
 23\%	& 10\%	&18\%	&22\%	&21\%	&19\% &19\% \\
\hline

\end{tabular}
\caption{Performance benefit due to software guidance}
\label{tab:avg-ones}
\end{table}

\subsection{The Impact of Software}
\label{sec:evalsoft}
All \PRAS configurations studied thus far, used software provided per layer precisions to reduce essential bit content. \PRAS does not require these precisions to operate. Table~\ref{tab:avg-ones} shows what fraction of the performance benefits is due to the software guidance for $\PRAS_{2b}^{1R}$, the best configuration studied. The results demonstrate that: 1)~\PRAS would outperform the other architectures even without software guidance, and 2)~on average, software guidance improves performance by 19\% which is on par with the estimate of Section~\ref{sec:motivation} for ideal \PRAS (from 10\% to 8\%). 


\subsection{Quantization}
\label{sec:quantized}

\begin{figure}[t]
\centering
\includegraphics[width=0.45\textwidth]{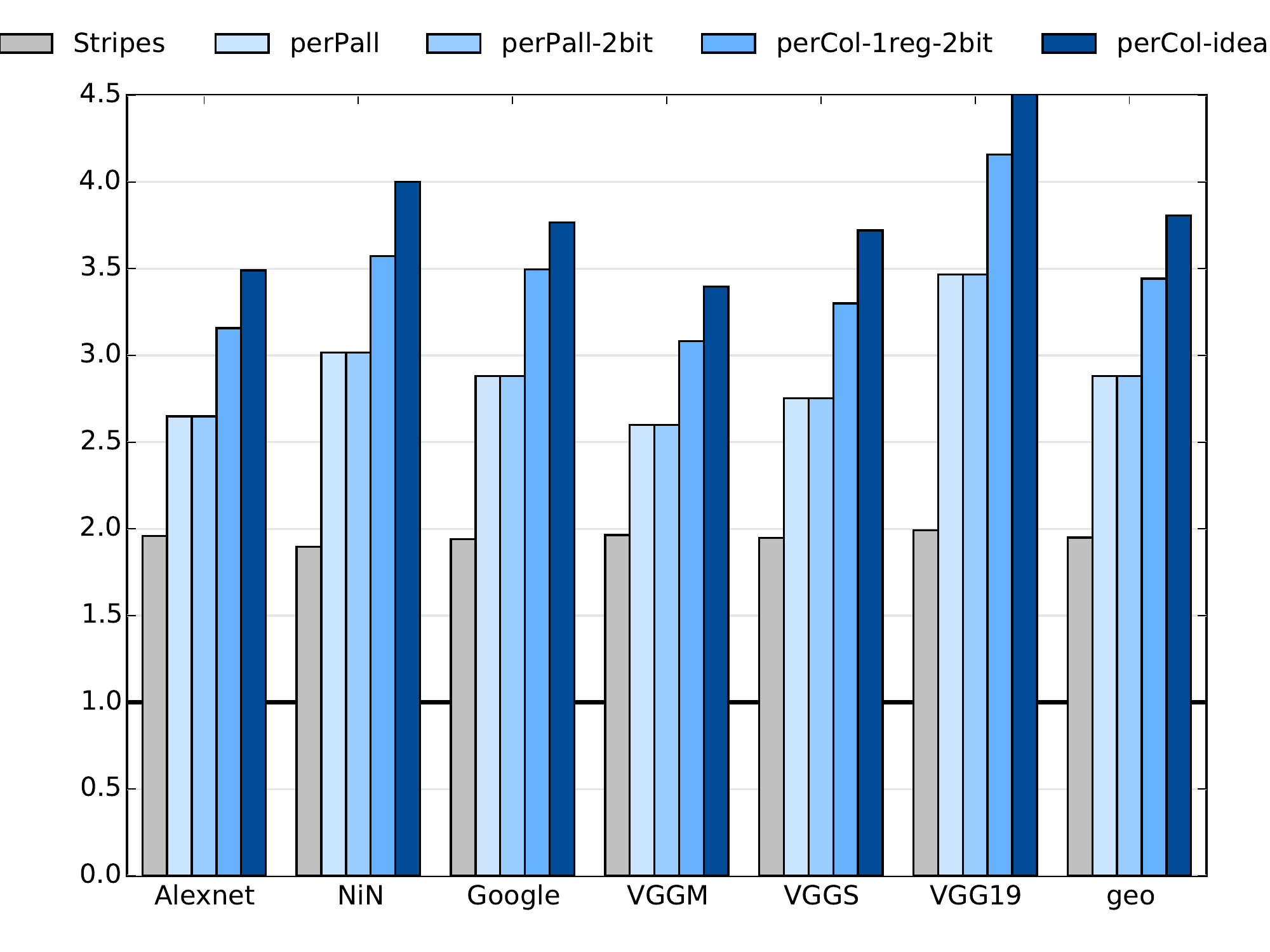}
\caption{Performance: 8-bit quantized representation.}
\label{fig:quantization}
\end{figure}

Figure~\ref{fig:quantization} reports performance for \BASE\ and \PRAS configurations using the 8-bit quantized representation used in Tensorflow~\cite{quantizedBlog, gemmlowp}. 
This quantization uses 8 bits to specify arbitrary minimum and maximum limits per layer for the neurons and the synapses separately, and maps the 256 available 8-bit values linearly into the resulting interval. This representation has higher flexibility and better utilization than the reduced precision approach of \STRL since the range doesn’t have to be symmetrical and the limits don’t have to be powers of two, while still allowing straightforward multiplication of the values.
The limit values are set to the maximum and the minimum neuron values for each layer and the quantization uses the recommended rounding mode.

Figure~\ref{fig:quantization} reports performance relative to \BASE for $\PRAS_{single}$, $\PRAS_{2b}$, $\PRAS_{2b}^{1R}$, and $\PRAS_{2b}^{\infty R}$. \PRAS  performance benefits persist and are nearly $3.5\times$ for $\PRAS_{2b}^{1R}$. Measuring the area and energy of these designs is left for future work, however, the absolute area and energy needed by all will be lower due to the narrower representation. Moreover, given that the tile logic will occupy relatively less area for the whole chip and given that the SB and NM account for significant area and energy, the overall overheads of the \PRAS designs over \BASE will be lower than that measured for the 16-bit fixed-point configurations.
\section{Related Work}
\label{sec:related}
The acceleration of Deep Learning is an active area of research and has yielded numerous proposals for hardware acceleration. \textit{DaDianNao} (\BASE) is the de facto standard for high-performance DNN acceleration~\cite{DaDiannao}. In the interest of space, this section restricts attention to methods that are either directly related to \BASE, or that follow a value-based  approach to DNN acceleration, as \PRA falls under this category of accelerators.
Value-based accelerators exploit the properties of the values being processed to further improve performance or energy beyond what is possible by exploiting computation structure alone. Cnvlutin~\cite{albericio:cnvlutin} and \STRL~\cite{Stripes-CAL}\cite{Stripes-MICRO} are such accelerators and they have been already discussed and compared against in this work.



\textit{PuDianNao} is a hardware accelerator that supports seven machine learning algorithms including DNNs~\cite{liu_pudiannao:_2015}. \textit{ShiDianNao} is a camera-integrated low power accelerator that exploits integration to reduce communication overheads and to further improve energy efficiency~\cite{du_shidiannao:_2015}. Cambricon is the first instruction set architecture for Deep Learning~\cite{cambricon:2016}.
Minerva is a highly automated software and hardware co-design approach targeting ultra low-voltage, highly-efficient DNN accelerators~\cite{Reagen2016}. 
Eyeriss is a low power, real-time DNN accelerator that exploits zero valued neurons for memory compression and energy reduction~\cite{isscc_2016_chen_eyeriss}. 
The Efficient Inference Engine (EIE) exploits efficient neuron and synapse representations and pruning to greatly reduce communication costs, to improve energy efficiency and to boost performance by avoiding certain ineffectual computations~\cite{han_eie:_2016}\cite{han_deep_2015-1}.  EIE targets fully-connected (FC) layers and was shown to be  $12\times$ more efficient than \BASE on FC layers, and $2\times$ less efficient for convolutional layers. All aforementioned accelerators use bit-parallel units. While this work has demonstrated \PRA as a modification of \BASE, its computation units and potentially, its general approach could be compatible with all aforementioned accelerator designs. This investigation is interesting future work.
As newer network architectures like GoogLeNet, NiN and VGG19 rely less on fully connected layers, this work used \BASE as an energy efficient and high performance baseline. 




Profiling has been used to determine the  precision requirements of a neural network  for a hardwired implementation~\cite{kim_x1000_2014}. EoP has been exploited in general purpose hardware and other application domains. For example, Brooks \textit{et al.}~\cite{Brooks:1999} exploit the prefix bits due to EoP to turn off parts of the datapath improving energy. Park \textit{et al.}~\cite{park_dynamic_2010}, use a similar approach to trade off image quality for improved energy efficiency. Neither approach directly improves performance. 





\section{Conclusion}
\label{sec:conclu}
To the best of our knowledge \PRA is the first DNN accelerator that exploits not only the per layer precision requirements of DNNs but also the essential bit information content of the neuron values. While this work targeted high-performance implementations, \PRA's core approach should be applicable to other hardware accelerators.

\bibliographystyle{ieeetr}
\bibliography{ref}

\end{document}